\documentclass[11pt]{article}
\usepackage{graphicx}
\usepackage{amsmath}
\usepackage{hyperref}
\usepackage{geometry}
\usepackage[numbers]{natbib}

\title{Forecasting  S\&P 500 Using LSTM Models}
\author{Prashant Pilla, Raji Mekonen}
\date{01/29/2025}

\begin{document}

\maketitle

\setlength{\parskip}{\baselineskip}

\begin{abstract}
\setlength{\parskip}{\baselineskip}
With the volatile, complex nature of financial data which is also influenced by many external factors, forecasting the stock market has been seen to be a challenging task. Traditional models like ARIMA and GARCH were observed to be good with linear data. However, the stock market data involves non-linear dependencies and intricate patterns that are better handled by machine learning and deep learning approaches. Taking that a step further to patch hyper-parameter tuning and computational complexity that machine learning lacks, we get deep learning models like Long Short-Term Memory (LSTM) networks. 
In this report, we compare ARIMA and LSTM models in predicting the S\&P 500 index, one of the most popular financial benchmarks. Using historical price data and technical indicators, we evaluated these models using the Mean Absolute Error (MAE) and Root Mean Squared Error (RMSE) metrics. The ARIMA model showcased reasonable performance with an MAE of 462.1, RMSE of 614, and an accuracy of 89.8\%. This demonstrated its effectiveness in capturing short-term trends but also showed that it is limited by its linear assumptions. The LSTM model, with favorable features, achieved an MAE of 369.32, RMSE of 412.84, and an accuracy of 92. 46\%, capturing both short- and long-term dependencies. The LSTM model without features outperformed the version with all features, achieving an MAE of 175.9, RMSE of 207.34, and an accuracy of 96.41\%, which showcased its ability to handle market data.
Accurately forecasting the stock market is crucial because of its effect on investment strategies, risk assessments, and market stability. By taking advantage of the sequential processing capabilities of LSTM, this report confirms how deep learning methods can handle volatile market conditions when compared to traditional models. The results of our analysis not only reaffirm the transformative potential of LSTM but also provide steps that can be taken to improve upon the model. Through this comprehensive study forecasting financial data, we aim to showcase the insights, limitations, and potential for prediction accuracy. 

\end{abstract}

\newpage

\section{Introduction}

Stock price forecasting has always been a fundamental and challenging problem when dealing with financial time series. When it comes to financial market data, there are many factors in play such as high volatility, non-linear dynamics, and sensitivity to many factors including historical prices, trading volumes, macroeconomic indicators, and investor sentiment. Although predicting exactly where stock prices may move is considered impossible, there are many tools that help investors when trying to forecast a market. These tools allow them to find trends, patterns, and potential price movements in order to have well-considered choices. However, the complexity of stock prices and many other factors can make this a very hard problem.  

Traditional time series models such as Autoregressive Integrated Moving Average (ARIMA) and Generalized Autoregressive Conditional Heteroskedasticity (GARCH) are a good base when trying to solve the challenging problem of forecasting financial data. This is because of their ability to model linear relationships and short-term patterns. ARIMA has been shown to be useful for finding trends and seasonality, while GARCH is better suited for modeling time-varying volatility. Their strengths come with some drawbacks with certain time series data because the models assume stationarity and linearity within the data. These limitations include handling non-linear dependencies, uncovering complex patterns over time, and finding long-term relationships within the data it is given. These limitations are the reason why it is less effective with financial data as it is non-linear, has some long-range dependencies, and is influenced by many factors.

Some of these challenges can be addressed by using machine/deep learning techniques, which have been shown to be better alternatives. Basic Recurrent Neural Networks (RNNs) introduced the ability to process sequential data by maintaining a hidden state that captures information from prior time steps. This allows the model to learn the temporal dependencies within the data for later use. However, RNNs face a critical limitation: they struggle to keep the information over long sequences because of the vanishing gradient problem. This happens when gradients used in the neural network become increasingly small during backpropagation, which ultimately hinders the network's ability to learn long-term dependencies. 

To go a step further, Long Short-Term Memory (LSTM) networks were developed to overcome the drawbacks of basic RNNs. LSTMs have a unique architecture that includes input, forget, and output gates. The memory capabilities of LSTM networks make it well-suited for financial time series forecasting, allowing the model to capture both short and long-term trends by keeping or discarding certain information. There are also LSTM variants such as Bidirectional LSTM (BiLSTM), Gated Recurrent Units (GRU), and Attention-LSTM that further improve the performance by enhancing parts of the model. These features allow the model to be very well-rounded and suitable for financial data. 

With constant change and many moving pieces in the stock market, it may be challenging for some investors to stay on top of a never-ending cycle of fluctuations. Being able to predict SPX prices can help aid investors in making the right decision and give insight into a never-ending stream of data. This paper explores the applications of machine learning models such as LSTM and traditional models such as ARIMA to forecast the S\&P 500 Index (SPX). The goal of using data such as historical prices and other financial metrics is to see if any underlying patterns in the data could help give insight into the market flow. With a combined interest in computer science and the financial markets, this project allows us to intersect the two fields. Also, diving deep into machine learning and AI, this project will help develop skills that can be used later on. 

\section{Literature Review}

\subsection{Traditional Statistical Methods for Stock Prediction}

Traditional statistical methods have been shown to be great methods for time series forecasting due to their simplicity, interpretability, and ability to perform well on smaller data sets. These models, however, have drawbacks when they are applied to financial data as it is highly complex and nonlinear. 

\subsubsection{ARIMA Model}
The Autoregressive Integrated Moving Average (ARIMA) model is highly used as a statistical approach for forecasting sequential data. As demonstrated by Adebiyi et al. \cite{paper4}, the ARIMA model can capture short-term dependencies and generate a reasonable prediction, which was used to forecast financial data. 

The ARIMA model consists of three parts:
\begin{itemize}
    \item \text{Autoregressive (AR)}: The linear relationship between its current value and its lagged observations.
    \item \text{Differencing (I)}: The number of times the series is differenced to achieve stationarity to remove trends and seasonality.
    \item \text{Moving Average (MA)}: Relates the current value to past forecast errors using a number of lagged errors.
\end{itemize}

The study by Adebiyi et al. applied ARIMA to predict daily stock prices and found that the model performed well on stationary time series. Their experiments involved rigorous parameter selection (using $p$, $d$, and $q$ terms) and model validation. They also acknowledged that ARIMA struggles with capturing the nonlinear and long-term dependencies with the given data set, which are rooted in the stock price's movements. 

\subsubsection{ARCH/GARCH Models}
While ARIMA focuses on the average behavior seen throughout the time series data, factoring in volatility is essential for forecasting financial data as it factors into the prices of the stock market. Autoregressive Conditional Heteroskedasticity (ARCH) and its generalized form Generalized ARCH (GARCH) models address this by explicitly modeling for time-varying volatility.

Raheem et al. \cite{paper11} explored ARCH/GARCH models to model volatility in financial time series. These models assume that variance at a given time depends on the squared residual errors of previous periods (ARCH) or combines past residuals with previous conditional variances (GARCH). Their model showed effectiveness in capturing volatility clusters, where it was seen that large changes in the stock price were followed by even larger changes, and small changes tend to follow small changes. 

The mathematical form of a GARCH(1,1) model can be represented as:
\[
\sigma_t^2 = \alpha_0 + \alpha_1 \epsilon_{t-1}^2 + \beta_1 \sigma_{t-1}^2
\]
where $\alpha$ is the ARCH parameter indicating how sensitive today's conditional volatility, $\sigma_t^2$, to the prior day's squared residual $\epsilon_{t-1}^2$. $\beta$ is the GARCH parameter, showing the persistence of the past volatility to its current $\sigma_t^2$.

Raheem et al. \cite{paper11} applied GARCH to stock market data and showed that it has strengths in forecasting volatility, which is crucial for pricing options and risk management. They also noted that the models are sensitive to outliers and they assume a symmetric response to shocks (positive or negative impacts), which is not always the case for the financial market.



\subsection{Machine Learning Approaches for Stock Prediction}

Machine learning approaches have been shown to be better alternatives to traditional statistical methods for stock prediction because they are able to model the complex, nonlinear relationships that can be found in financial time series data. Unlike traditional models, machine learning algorithms can analyze larger data sets, identify patterns, and adapt to changes made in the data over time. This makes them well suited for the dynamic and uncertain nature seen in the stock market. 

\subsubsection{Support Vector Regression (SVR) with Grey Correlation Degree}
Support Vector Regression (SVR) is a machine learning method that is popular for its use in handling high dimensional and nonlinear data. Wang \cite{paper2} showed how this model combined with Grey Correlation Degree (GCD) can be used to enhance accuracy by optimizing feature selection which improved stock price forecasting. GCD assigns weights to input features based on their correlation with its target variable. This allows SVR to focus on the most relevant features in order to make its prediction.

SVR is comprised of a hyperplane in a high-dimensional space which minimizes the error bet between the predicted and actual values while keeping the model as generalized as possible. By introducing GCD, enhancements are made to SVR by reducing noise and focusing solely on the critical predictors which addresses the key challenge in stock prediction: finding the most impactful features. By improving the feature selection, Wang \cite{paper2} showed a significant boost in accuracy when applying it to the dataset. This preprocessing technique showed how well a model can perform given the right features it trains on. 

\subsubsection{k-Nearest Neighbors (KNN)}
The k-Nearest Neighbors (KNN) is an algorithm that predicts future values by finding past values that are most closely related to the current one. Alkhatib et al. \cite{paper10} used the KNN algorithm to forecast stock prices for six major companies listed on the Jordanian Stock Exchange. The KNN algorithm measures the Euclidean distance between data points to identify the closest matches and uses these neighbors to make predictions. KNN does not make assumptions on the underlying data showing that it is a nonparametric method. Alkhatib et al. \cite{paper10} showed that KNN was quite effective when predicting closing prices. This is because of its simple structure and its ability to adapt to nonlinear relationships within the data. During this study, the evaluation of KNN's robustness for stock price prediction by measuring error ratios was seen to be low. The authors highlighted that selecting the right amount of neighbors and distance metrics is key for performing well. It is also worth noting that this model does struggle with very large datasets and noisy data if the right preprocessing is not performed.

\subsubsection{Random Forest for Stock Price Prediction}

Random Forest is a machine learning method that combines multiple decision trees to improve prediction accuracy and reduce overfitting. Khaidem et al. \cite{paper12} applied Random Forest to predict the direction of stock market prices using historical stock data and technical indicators to forecast stock for major companies like Apple (AAPL) and General Electric (GE). Random Forest generates several decision trees, each trained on random subsets of features and data points. The predictions from individual trees are combined by averaging to produce a final output. Khaidem et al. \cite{paper12} showed this method achieves higher prediction accuracy due to its ability to model nonlinear patterns in financial data. 

\subsubsection{Hidden Markov Models (HMM)}
Hidden Markov Models (HMM) are used widely for time series forecasting to model systems that transition between hidden states over time. Hasanbas \cite{paper7} applied HMM to forecast financial time series to show its ability to capture the probabilistic states of stock market trends. In this application of the model, HMM assumes that the observed time series data are generated by a sequence of hidden states, each of which each follows a distinct probability distribution. 

HMM consists of three parts:
\begin{itemize}
    \item \text{Hidden States}: Unobserved states that the system transitions between over time
    \item \text{Transition Probabilities}: Probabilities of transitioning from one hidden state to another at each time step.
    \item \text{Emission Probabilities}: Probabilities of observing a particular output given the hidden state.
\end{itemize}

For stock prediction, HMM identifies market regimes (e.g., bullish, bearish, or stable) by analyzing transitions between these states based on historical price movements. Hasanbas \cite{paper7} showed that HMM effectively captures short-term dependencies, making it well-suited for a volatile stock market. It is important to note that HMM assumes Markovian properties, meaning its probabilities of transitioning to the next state only depend on its current state. This makes it less viable for forecasting long-term. 

\subsubsection{Bayesian Time Series Analysis}
The Bayesian method is another probabilistic method used for time series forecasting. Unlike the previous model, the Bayesian model introduces uncertainty into the predictions. Steel \cite{paper8} showed the application of Bayesian time series analysis for stock prediction. This model involves constructing posterior distributions for model parameters using Bayes' theorem. For financial data, this model is well suited as it can handle noisy, incomplete, or volatile data. Steel \cite{paper8} highlighted that Bayesian methods can outperform deterministic approaches in volatile markets by accounting for model and parameter uncertainties.



\subsection{Deep Learning Methods}

Deep learning methods have been seen to be an improvement upon traditional statistical and machine learning methods when it comes to very large complex data sets. These models excel at capturing long-term dependencies, modeling complex relationships, and being able to adapt to high-dimensional financial time series data. 

\subsubsection{Long Short-Term Memory (LSTM) Networks and Recurrent Neural Networks (RNNs)}

Long Short-Term Memory (LSTM) networks have been shown to be one of the best choices when it comes to handling sequential data. They are an improvement upon RNNs, which struggle with the vanishing gradient problem, where information from early time steps tends to fade away. Their unique architecture comprises memory cells along with three gates:
\begin{itemize}
    \item \text{Input Gates}: Controls which parts of the previous memory are discarded.
    \item \text{Forget Gates}: Determines what new information should be added to the memory.
    \item \text{Output Gates}: Regulates what information is output for the current step.
\end{itemize}
These components work together to selectively retain, update, and output information to capture both the short-term fluctuations and long-term trends in the dataset. Zou and Qu \cite{paper1} implemented LSTM networks to predict stock prices and analyze this strategy against others. Their study showed that the LSTM model outperformed the traditional RNNs and machine learning methods by effectively learning patterns in historical stock price data. Sonkavde et al. \cite{paper3} conducted a review comparing machine learning and deep learning methods which included LSTM, Convolutional Neural Networks (CNNs), and traditional RNNs. Their result showed the effectiveness of the LSTM model compared to RNNs and CNNs by handling temporal and spatial features better. 

\subsubsection{Alpha-RNN}
Another improved variant of RNNs is called \(\alpha\)-RNN, which offers some computational efficiency while being able to maintain good prediction accuracy. Dixon and London \cite{paper9} proposed using \(\alpha\)-RNN for forecasting financial time series data. The \(\alpha\)-RNN enhances traditional RNNs by introducing an exponential smoothing layer which helps the model retain long-term information in the data by combining past and current states. Dixon and London showed that \(\alpha\)-RNNs are competitive in financial forecasting as they offer a reduced computational overhead when compared to LSTM. However, they lack fine-grained control over information flow which allows LSTM to handle complex, long-term dependencies. 

\subsubsection{Graph Neural Networks (GNNs) for Financial Time Series Prediction}

Exploring spatial and relational information in financial data has led some researchers to use Graph Neural Networks (GNNs). GNNs are designed to process data that is represented as graphs, where each node and edge encode entities and their relationships.
Xiang et al. \cite{paper6} proposed using a variant of GNNs called Temporal and Heterogenous GNN (TH-GNN). This allowed them to try and forecast financial time series by modeling a relationship between several financial indicators and the market data. Unlike the sequential model LSTM, this model captures dependencies between different entities, such as different stocks and indices, over time providing a better understanding of the market. 

\subsubsection{Federated Learning with Large Language Models (LLMs)}

Another emerging trend in deep learning is integrating Federated Learning (FL) and Large Language Models (LLMs) for financial forecasting. FL is a decentralized machine learning technique that allows multiple edge devices (clients) to collaboratively train a global model while keeping the data localized. LLMs are advanced machine-learning models designed to understand and generate human-like text. They are built using the transformer architecture, where text is converted into numerical representations called tokens, which are then processed through a series of encoder and/or decoder layers that analyze relationships between words in a sentence. Abdel-Sater and Hamza \cite{paper5} introduced an FL model which incorporated a pre-trained LLM. Instead of processing text, Abdel-Sater and Hamza adapted the LLM to handle stock data. This allowed them to use the collaborative training on multiple data sources without sharing any raw financial data. However, this comes with the cost of being less efficient on numerical financial data as opposed to the LSTM model. 



\subsection{Reflection}

Stock price forecasting is a challenging task that has evolved from using traditional statistical methods, such as ARIMA and GARCH, to more advanced machine learning and deep learning methods. Although machine learning methods improved the handling of non-linear data, as seen in SVR, KNN, and Random Forest, they have drawbacks in computational efficiency and handling high-dimensional data, such as time series. These drawbacks can be overcome by using Deep Learning methods such as LSTM, \(\alpha\)-RNNs, Graph Neural Networks (GNNs), and Federated Learning with LLMs. 
LSTM networks above all have emerged to be a leading solution by effectively capturing short and long-term dependencies in financial time series data while being computationally efficient.



\section{Approach to Solve the Problem}
This study compares two models, ARIMA and LSTM, to forecast SPX. Both models leverage their own unique methodologies to forecast time series data, and by comparing both models, it will highlight their respective strengths and weaknesses. 

\subsection{Data Sources and Features}

\noindent The dataset used in this study encompasses daily values for the S\&P 500 (SPX) over the period of October 2013 to September 2024. The historical SPX data include:

\begin{itemize}
    \item 50 and 200-day Moving Averages (MOV\_AVG\_50/200D)
    \item 14-day Relative Strength Index (RSI\_14D)
    \item Open and Closing Prices (PX\_OPEN/CLOSE)
    \item High and Low Prices (PX\_HIGH/LOW)
    \item Daily Price High-Low Difference (PX\_HIGH\_LOW\_DIFFERENCE)
    \item Daily Volume (PX\_VOLUME)
    \item 30-day Volatility (VOLATILITY\_30D)
    \item Beta (BETA\_ADJ\_OVERRIDABLE)
\end{itemize}

\noindent Along with the historical SPX data, the following additional metrics are considered:

\begin{itemize}
    \item \text{SPX Ratios:}
    \begin{itemize}
        \item Price-to-Earnings Ratio (PE\_RATIO)
        \item Price-to-Book Ratio (PX\_TO\_BOOK\_RATIO)
        \item Price-to-Sales Ratio (PX\_TO\_SALES\_RATIO)
        \item Earnings Yield (EARN\_YLD)
    \end{itemize}

    \item \text{Market Metrics:}
    \begin{itemize}
        \item Volatility Index (VIX)
        \item 10-Year Treasury Yield (USGG10YR)
        \item NAPM Manufacturing PMI (NAPMPMI)
        \item Consumer Confidence Index (CONCCONF)
    \end{itemize}
\end{itemize}

\noindent These metrics were chosen as they provide a comprehensive overview of the market and economic environment. All of the data was gathered from Bloomberg.

\subsection{Models}
\noindent The ARIMA model is a statistical time-series model that incorporates the three components AutoRegressive (AR), Integrated (I), and Moving Average (MA) for the parameters ($p, d, q$). This model requires the data to be stationary where the properties mean, variance, and autocorrelation should remain constant over time. Therefore, taking the difference from the current time step to the next one is done to remove trends. For this study, Auto-ARIMA is used to automate the selection of the optimal parameters ($p, d, q$) for the model.

\noindent The LSTM model is a deep learning model designed for sequential data. The network processes an input sequence, maintaining memory through three gates; forget, input, and output. For this study, the architecture of the neural network uses two LSTM layers with 64 neurons each with dropout layers to prevent overfitting by dropping a random 20\% of neurons after each iteration. Also, a dense output layer will be used to predict the next price with the Adam optimizer and Mean Squared Error (MSE) as the loss function.

\section{Experiment Design}
\subsection{Data Preprocessing}
With all the features obtained, key features were selected based on their correlation with SPX closing price (Figure 1), through a correlation vector. Features that showed strong positive or negative correlations were prioritized, with absolute values greater than 0.5, ensuring the LSTM model used relevant predictive information. For the ARIMA model, only the SPX closing price was used as it only works with just one time series and doesn't take other features into account.

\noindent Data Splitting:
\begin{itemize}
\item For ARIMA: The dataset is split into \text{80\% training} to fit the ARIMA model capturing ($p, d, q$), and \text{20\% testing} to validate predictions.
2\item For LSTM: The dataset is split into \text{60\% training}, \text{20\% validation}, and \text{20\% testing} all in chronological order. The sliding window technique is applied to generate feature-target pairs for LSTM. This method is used in time series forecasting where fixed-size windows of past values are used as inputs (features) to predict the next value (target). A window size of 216 was used with the target being 1 day out for both the training and testing set. Two separate models will assess the forecast with one including the favorable features and one without. 
\end{itemize}
\noindent Forward Filling:
\begin{itemize}
    \item Ensures a complete dataset by filling in missing values for features that are not set daily.
    \item Provides consistent values for all features at each time step.
\end{itemize}
\noindent Normalization:
\begin{itemize}
    \item All features are scaled using \textit{MinMaxScaler} to make sure they are in the same range (0-1). This prevents outliers in the data set from dominating the model. The features were scaled after splitting the data to ensure integrity (the model won't train on unseen data).
\end{itemize}

\begin{figure}[h!]
    \centering
    \includegraphics[width=.85\textwidth]{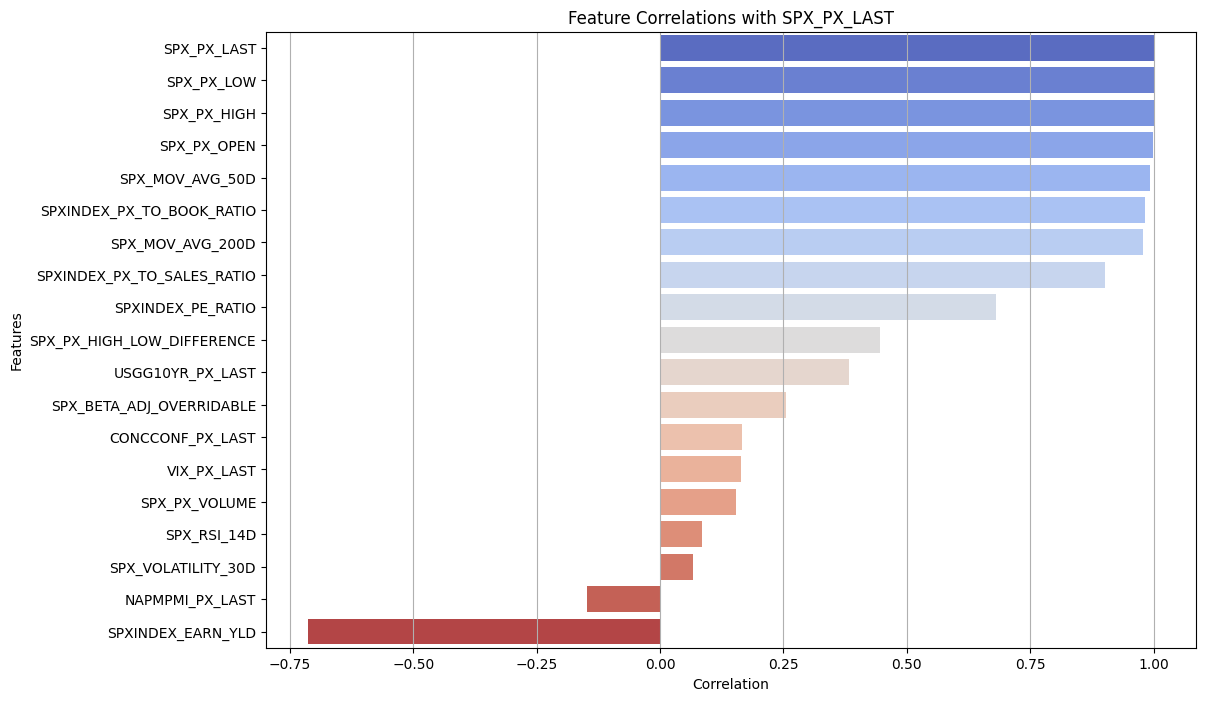}
    \caption{Price Correlation with Features}
    \label{fig:stock_chart}
\end{figure}
\subsection{Evaluation Metrics}
To compare the models, the following metrics are used: 
\begin{itemize}
    \item \text{Mean Absolute Error (MAE)}: Measures the average magnitude of the errors in a set of predictions by taking the sum of the absolute difference between the actual and predicted values over a number of observations. 
    \item \text{Root Mean Squared Error (RMSE)}: Measures the square root of the average of squared differences between actual and predicted values. 
    \item \text{Accuracy}: Shows how close predicted values are to actual with the equation: $$\text{Accuracy} = 100 - \left( \frac{\text{MAE}}{\text{Mean of Actual Values}} \times 100 \right)$$
\end{itemize}

\section{Experiment Results}
\subsection{ARIMA}
The ARIMA model showed a reasonable performance with the results showing an MAE value of 462.1 and RMSE value of 614. The results showed an accuracy percentage of 89.8\%. The results (Figure 2) showed this model to be effective in capturing short-term trends in the data, but its reliance on linear assumptions limited its ability to capture the entire testing set. 

\begin{figure}[h!]
    \centering
    \includegraphics[width=.85\textwidth]{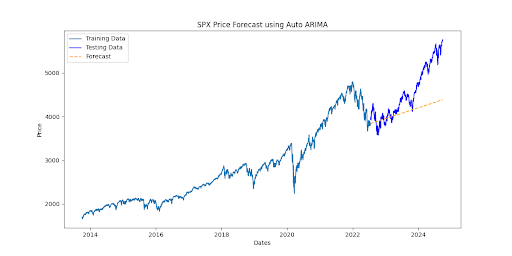}
    \caption{ARIMA model}
    \label{fig:stock_chart}
\end{figure}

\subsection{LSTM}
The LSTM model performed well, outperforming the ARIMA model in both forecasts. For the model that included favorable features (Figure 3), it achieved an MAE of 369.32 and an RMSE of 412.84. This showed an accuracy of 92.46\%, showing it captured both the short-term and long-term dependencies. The model that did not include features (Figure 4) showed an MAE of 175.9 with an RMSE being 207.34. This gave the model an accuracy of 96.41\%, showing a better performance than the LSTM model that did include features. It is important to note that there were minor discrepancies seen at highly volatile periods, which suggests potential improvements in the model's arguments or doing more feature engineering. 

\begin{figure}[h!]
    \centering
    \includegraphics[width=.85\textwidth]{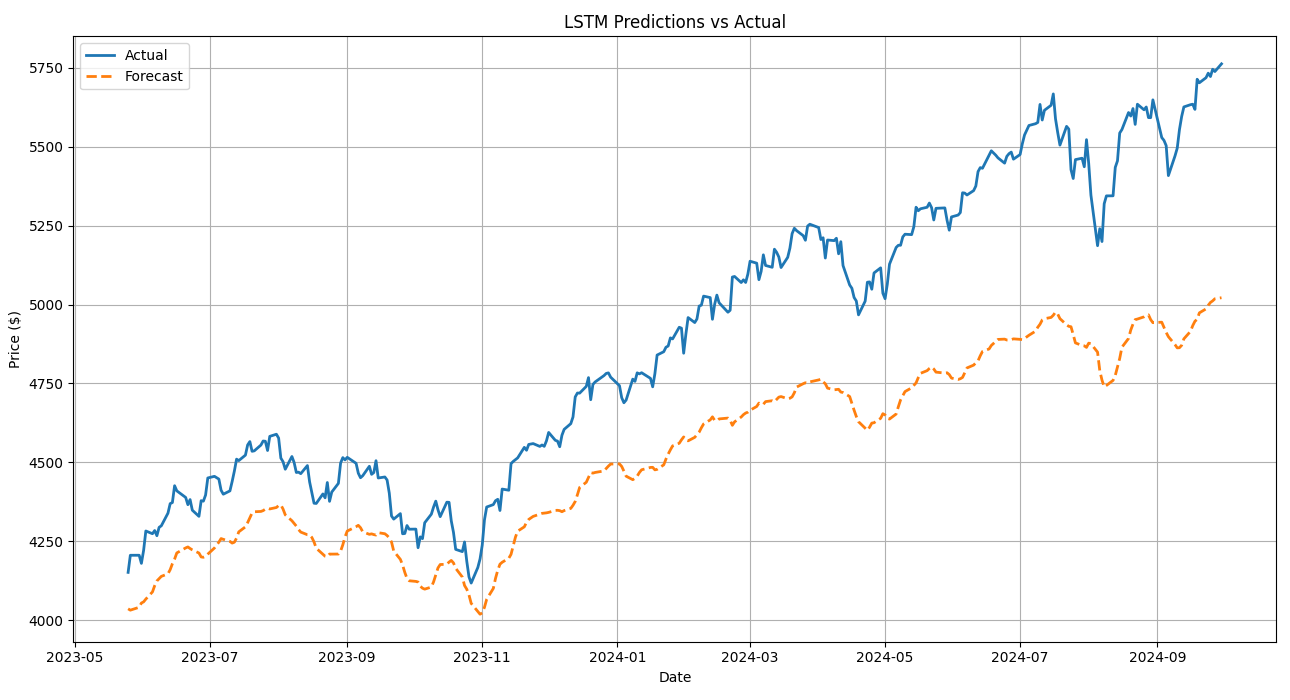}
    \caption{LSTM model with features}
    \label{fig:stock_chart}
\end{figure}
\begin{figure}[h!]
    \centering
    \includegraphics[width=.85\textwidth]{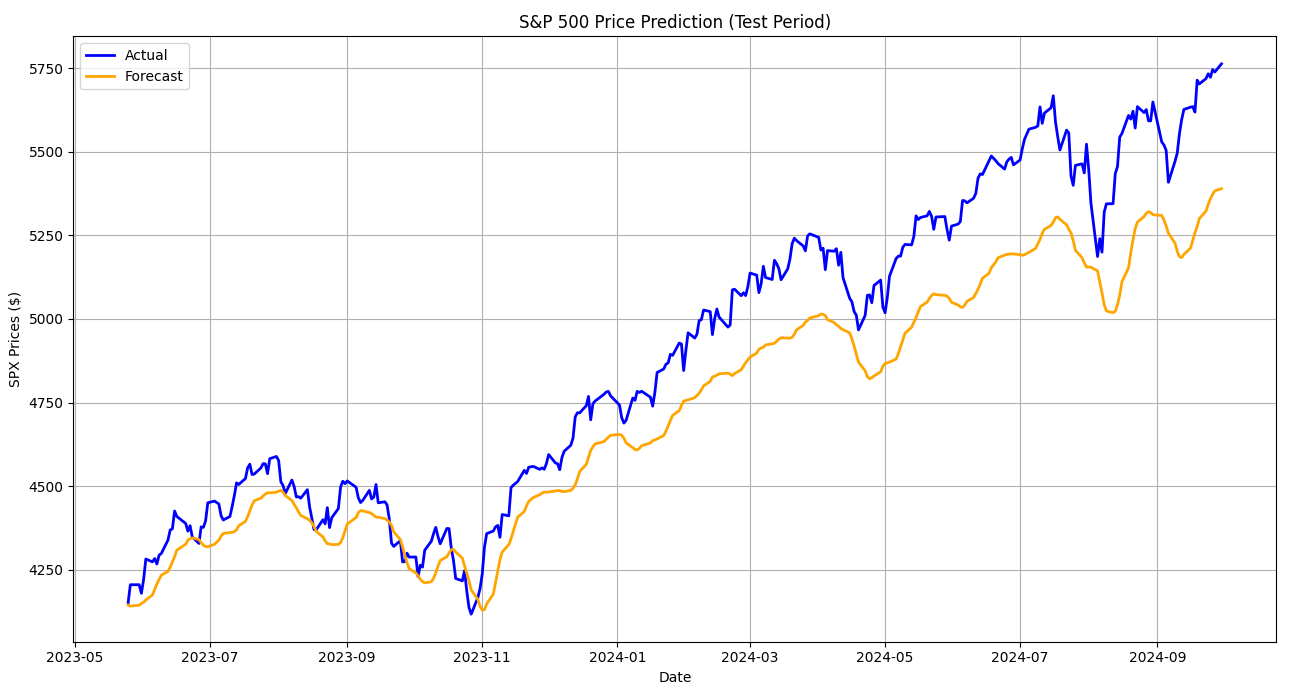}
    \caption{LSTM model without features}
    \label{fig:stock_chart}
\end{figure}

\section{Analysis of Results}
The results from the ARIMA model showcased the performance of the statistical time series forecast. From the evaluation metrics, it can be seen that the model captured 89.8\% of the variability in the SPX prices during the test period. This came from the MAE score of 462.1, meaning on average the ARIMA model's predictions were off by 462.1. It can be observed that the strong performance is seen mainly at the start of the test stage where it achieved 95.84\% accuracy in the first half, showing it can be a good tool to be used for short-term trends. However, the accuracy gradually declined with an accuracy of 84.92\% in the later half of the forecast. This decline indicates that while the ARIMA model can show effectiveness for short-term forecasting, it may struggle to capture longer-term nonlinear trends in the dataset. With a RMSE value of 614, the model showed to have significant deviations in the prediction. In comparison to previous work, Adebiyi et al. \cite{paper4} showed the effectiveness of ARIMA for stock price forecasting but also acknowledged its limitations when dealing with nonlinear and highly volatile financial data.

The results for the LSTM model significantly outperformed ARIMA: with and without additional features. The results showed the model's architecture capturing both short and long-term trends. With an MAE score of 369.32 and 175.9, the models showed to make more accurate predictions when compared to the ARIMA model. Also with an RMSE score of 412.84 and 207.34 for both the models, it showed the models minimized larger deviations and were able to handle some of the volatile movements. The LSTM model without additional features notably outperformed the model with features. This can be attributed to several factors such as feature noise where additional features may introduce noise or redundant information that can hinder the model's performance. Also, the LSTM model is designed for sequential data, effectively capturing the underlying trends and dependencies in time series data without requiring additional features or indicators. From Figures 3 and 4, it can be seen that the predictions closely followed the actual SPX price, but there is a minor decline in the last quarter of the prediction due to it not realizing new dependencies that might have been made. Zou and Qu \cite{paper1} showed in their model, it outperformed traditional machine learning models and RNNs with financial data, attributing this to LSTM's ability to handle long-term dependencies and nonlinear relationships. This mirrors the result of this study where the LSTM achieved a 96.41\% accuracy without additional features. 

\section{Conclusion}
In this study, the application of two time series models, ARIMA and LSTM, were used to forecast the S\&P 500 (SPX) index. 

The ARIMA model served as a traditional statistical benchmark, showing how effective it is in capturing short-term trends but being limited in handling nonlinearity and long-term dependencies. Our results showed that ARIMA achieved an MAE of 462.1, RMSE of 614.0, and an accuracy of 89.8\%, with its performance declining over longer forecasts. 

In contrast, the LSTM model, which is a deep learning approach designed for sequential data, significantly outperformed the ARIMA model. The LSTM model without additional features achieved the best results with an MAE of 175.9, RMSE of 207.34, and an accuracy of 96.41\%. The model that included favorable features demonstrated slightly lower performance with an MAE of 369.32, RMSE of 412.84, and an accuracy of 92.46\%. This difference in performance also highlights the importance of good feature engineering and noise reduction. These results also align with some of the findings from prior literature such as Zou and Qu \cite{paper1} and Sonkavde et al. \cite{paper3}, where LSTM models were shown to effectively capture long and short-term dependencies in financial data. 

Although the LSTM model showed superior performance, there are still areas where it can be improved. Future work includes:
\begin{itemize}
    \item Hybrid Model Integration: Combining the strengths of this model with others such as GARCH for its ability to handle volatility or GNNs to capture the relationship of the financial data with other features. 
    \item Optimizing LSTM Architecture: Looking into ways to incorporate LSTM variants such as Bidirectional LSTM, Attention-LSTM, or incorporate a transformer-based model as seen by Abdel-Sater and Hamza \cite{paper5}. 
    \item Generalizing the Model: Seeing the model performed well on SPX data, testing this out on other indices or stocks may show its robustness and capabilities across markets. 
\end{itemize}

Through this study, we showed the potential of LSTM networks highlighting the powerful approach deep learning models are capable of. Our findings showed that the LSTM model offers a powerful way to handle volatile, nonlinear, complex data seen in financial markets, showing a direction to where stock prediction can go.

\end{document}